%% file: main.tex
\documentclass[runningheads]{llncs}
\usepackage[T1]{fontenc}
\usepackage{graphicx}
\usepackage[hidelinks]{hyperref}
\usepackage{amsmath,amsfonts,amssymb}
\usepackage{mathtools}
\usepackage{booktabs}
\usepackage{enumitem}
\usepackage{bm}
\usepackage{subcaption}
\usepackage{tabularx}
\usepackage{siunitx}
\usepackage{multirow}
\usepackage{multicol}
\usepackage{array}
\usepackage{arydshln}
\usepackage{wrapfig}
\usepackage{subcaption}
\usepackage{ragged2e}
\newcommand{\mypar}[1]{\smallskip\noindent\textbf{#1.}}
\newcommand{\mypartwo}[1]{\vspace{0.5pt}\noindent\textit{#1.}}

\usepackage{xcolor}

\begin{document}
%
\title{Mind the Long Tail: Understanding the Difficulty of Delay Detection in Business Processes}
%
\titlerunning{Understanding the Difficulty of Delay Detection in Business Processes}

%
\author{Keyvan Amiri Elyasi\thanks{Equal contribution}\inst{1} \orcidID{0009-0007-3016-2392} \and Lukas Kirchdorfer$^\star$\inst{1,2} \orcidID{0000-0003-4713-9328} \and Heiner Stuckenschmidt\inst{1} \orcidID{0000-0002-0209-3859}}
\authorrunning{K. Amiri Elyasi et al.}
\institute{Data and Web Science Group, University of Mannheim, Germany\\
\email{\{keyvan.amiri, heiner.stuckenschmidt\}@uni-mannheim.de} \and SAP Signavio, Walldorf, Germany\\ \email{lukas.kirchdorfer@sap.com}}
\maketitle              

\begin{abstract}
The early detection of delayed cases in business processes is a critical capability for organizations. Predictive process monitoring (PPM) supports this task by using historical event logs to predict the remaining time of ongoing cases, enabling timely interventions to avoid missed deadlines and service level violations. Although remaining time prediction has advanced considerably through sophisticated deep learning architectures, little is known about the intrinsic difficulty of delay detection itself. Since performance is typically assessed using aggregate metrics, prior work provides limited insight into how models perform across the target distribution, especially on the operationally most critical cases with large delays.
In this paper, we address this gap by analyzing the difficulty of delay detection. Across 14 event logs, we show that remaining times are typically strongly right-skewed, with only a small fraction of cases exhibiting large delays. Existing models capture the mode of this distribution well but perform poorly on high-delay cases. We further uncover pronounced heteroscedasticity, showing that predictive uncertainty increases with delay magnitude.
Based on these findings, we evaluate approaches to mitigate the imbalance problem, but find only limited benefits, suggesting that the key underlying problem may not be imbalance but higher uncertainty associated with delayed cases. We show that this correlation can be exploited to substantially improve the identification of delayed cases.
Overall, our work provides new insights into the sources of difficulty in delay detection and identifies uncertainty-aware modeling as a promising direction for future PPM research.

\end{abstract}

\input{Sections/01_Intro}

\input{Sections/02_Motivation}

\input{Sections/03_Setting}

\input{Sections/04_Imbalanced}

\input{Sections/05_uncertainty}

\input{Sections/06_Related_work}

\input{Sections/07_conclusion}

\bibliographystyle{splncs04}
\bibliography{bibliography}

\end{document}

%% file: Sections/01_Intro.tex
\section{Introduction}
\label{sec:intro}

Organizations rely on business processes to deliver services to customers and ensure the timely execution of work. In this context, execution delays are a major concern, as they lead to missed deadlines, violated service level agreements, increased operational costs, and diminished customer satisfaction~\cite{tater2018prediction,zeng2008using,kim2017early}. Consequently, the early identification of delay-prone cases is essential for effective process management. Such identification can leverage execution data recorded by enterprise information systems in the form of event logs, which document historical process executions and provide a basis for anticipating future behavior.

In business process management, this is addressed through predictive process monitoring (PPM), which aims to predict future characteristics of running cases based on historical event data~\cite{di2022predictive}. Among PPM tasks, remaining time prediction is one of the most widely studied. Given the observed prefix of an ongoing case, it estimates the time until completion and provides an actionable signal for identifying potentially delayed cases and enabling timely interventions.

In recent years, substantial progress has been made in remaining time prediction, largely driven by advances in deep learning architectures such as LSTMs, Transformers, and graph neural networks~\cite{navarin2017lstm,amiri2024pgtnet,hennig2025leveraging}. While these approaches improve predictive performance, they are typically evaluated using aggregate metrics such as mean absolute error (MAE), which obscure where models perform well or fail. In particular, such metrics do not capture performance on rare delayed cases, despite their critical operational impact due to their potential to cause service level violations and increased costs.

As we show in this paper, distributions of remaining time in business processes are typically highly right-skewed: most cases complete within an expected time frame, whereas a small number exhibit extreme delays. Accurately identifying these rare delayed cases is essential for effective process management, yet it remains unclear whether existing models can reliably detect them. More fundamentally, there is still limited understanding of the intrinsic difficulty of delay detection in business processes, which hinders the development of more effective PPM approaches grounded in process-specific insights rather than relying primarily on advances in general-purpose machine learning models.

To address these challenges, this paper makes three contributions.
\begin{enumerate}[nosep, leftmargin=1em]
    \item First, we study the intrinsic difficulty of delay detection in remaining time prediction. Our analysis shows that (a) remaining times typically follow a strongly right-skewed distribution, (b) existing approaches capture the mode of this distribution well but fail to predict cases with large delays accurately, and (c) predictive uncertainty is heteroscedastic and increases with remaining time, indicating that high-delay cases are substantially noisier than normal cases.    

    \item Second, building on findings (a) and (b), we hypothesize that a key reason why existing models fail on high-delay cases is the identified \textit{target imbalance}. We therefore evaluate established imbalanced regression approaches and find, however, that their impact is limited.

    \item Third, based on finding (c), we investigate the utility of heteroscedastic uncertainty estimates for delay detection. Specifically, we study whether the relationship between predictive uncertainty and remaining time magnitude can be exploited to improve the identification of delayed cases. Our results show that uncertainty-aware models can substantially improve delay detection.
    
\end{enumerate}
   
\noindent
The remainder of the paper is structured as follows. \autoref{sec:motivation} analyzes the difficulty of delay detection across a variety of processes. \autoref{sec:setting} describes the experimental setup, followed by the evaluation of imbalanced regression approaches in \autoref{sec:imbalanced} and the assessment of heteroscedastic uncertainty in \autoref{sec:uncertainty}. Finally, we discuss related work in \autoref{sec:related_work} and conclude the paper in \autoref{sec:conclusion}.

%% file: Sections/02_Motivation.tex
\section{Motivation and Analysis}
\label{sec:motivation}
We illustrate the difficulty of delay detection using the \textit{Prepaid Travel Cost} process from the BPIC20 benchmark and then show that the observed patterns generalize across other event logs.

\begin{figure}[htbp]
    \centering
    \includegraphics[width=0.9\linewidth]{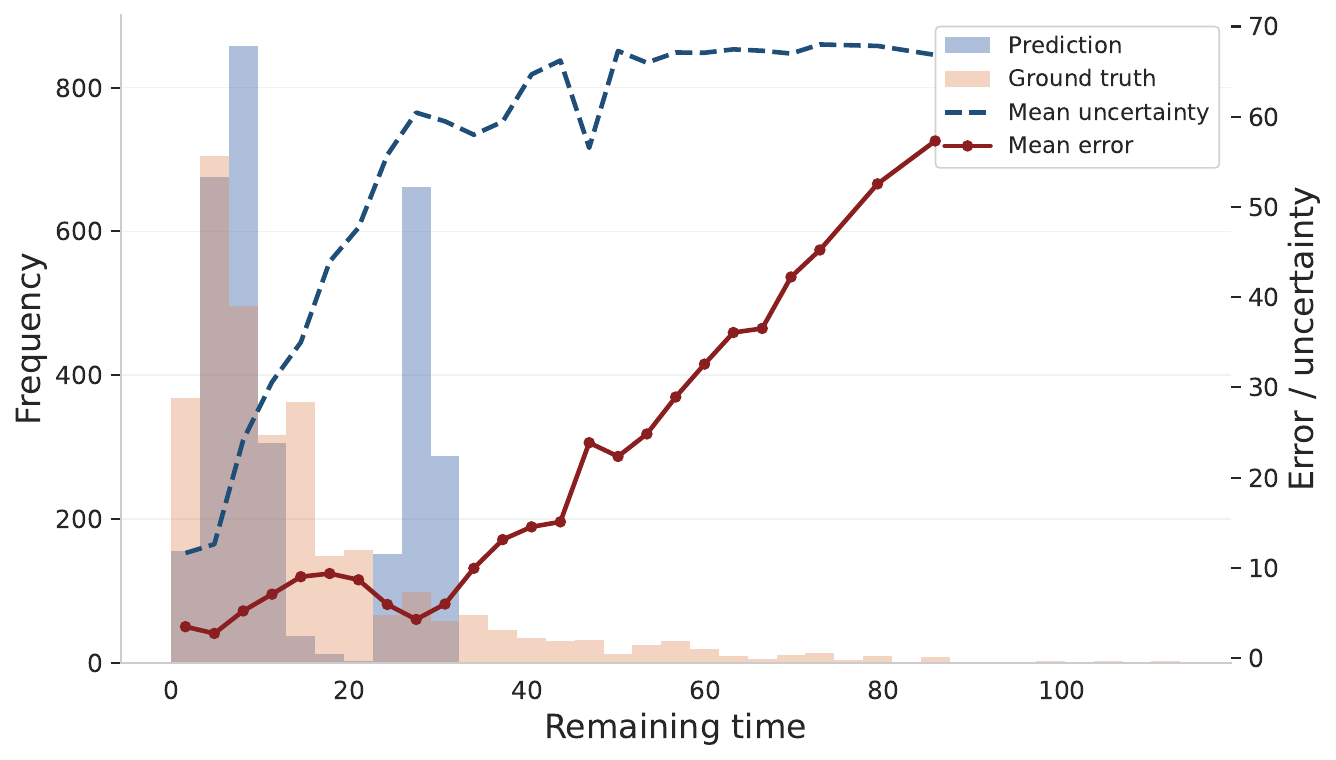}
    \caption{BPIC20PTC distribution of ground-truth and predicted remaining times (in days), with average error and heteroscedastic uncertainty per target range.}
    \label{fig:motivation}
 \end{figure}

\mypar{Illustration}
To illustrate the difficulty of delay detection, we analyze the BPIC20PTC process from Eindhoven University of Technology, which contains events related to two years of travel expense claims. \autoref{fig:motivation} shows the distribution of remaining times across all possible prefix positions in the process (orange histogram) and compares it with the distribution of predicted remaining times (blue histogram) produced by a standard LSTM-based approach~\cite{navarin2017lstm}.

The ground-truth distribution of remaining times is \textit{strongly right-skewed}, with most observations concentrated around 10 days and only a small number extending to 40--100 days. In contrast, the predictions are compressed into a much narrower range and fail to capture high-delay cases. The model never predicts a remaining time above 35 days. As a result, a case that actually requires, for example, 80 days until completion would receive a prediction of only 30--35 days. Such errors render the prediction ineffective for timely delay identification and operational intervention. Consequently, prediction error increases substantially toward the tail of the distribution, as illustrated by the red line.

Another key insight is that predictive uncertainty increases with remaining time, exhibiting a clear \textit{heteroscedastic} pattern. In other words, the figure suggests a correlation between the magnitude of uncertainty and the magnitude of remaining time.
Heteroscedastic uncertainty is data-dependent and reflects the irreducible noise inherent in the data generation process, and arises from stochastic and unobserved factors that influence process execution~\cite{weytjens_learning_2022}.
In our example process, the uncertainty is low around the mode of the distribution, where the majority of normal cases are located, but increases sharply for larger remaining times. It reaches its highest values around 40 days and remains elevated thereafter. 
This indicates that high-delay cases are not only rare, but also substantially noisier and therefore more difficult to predict than normal cases.

\mypar{Analysis Across Event Logs}
To assess whether these observations are the rule rather than the exception, we analyze 14 publicly available event logs, focusing on the skewness of the remaining time distribution and the degree of heteroscedastic uncertainty.

Table~\ref{tab_logs} summarizes the characteristics of the event logs. It reports the Fisher moment coefficient of skewness ($\gamma$) of the remaining time distribution, which quantifies its asymmetry. Positive skewness indicates a long right tail, with larger values corresponding to stronger asymmetry (for example, $\gamma = 2$ for an exponential distribution). Most logs exhibit pronounced positive skewness, indicating heavy-tailed remaining time distributions where most cases complete within a normal time frame, while a small number experience extremely long durations.

\input{Tables/event_logs}

\input{Tables/heteroscedastic}

To examine how prediction error and predictive uncertainty relate to remaining time, we analyze their association with the ground-truth remaining time $y$ on the test sets (final 20\% of cases) of each event log. Specifically, we compute the Spearman correlation between $y$ and two quantities: (i) the absolute prediction error $|\hat{y}-y|$ of a standard LSTM model trained with MAE~\cite{navarin2017lstm}, and (ii) the width of the prediction interval from an uncertainty-aware model based on a survival formulation~\cite{george2014survival}. Details on this model are provided in \autoref{sec:uncertainty}; here, it suffices to note that, unlike the standard LSTM, it produces both point estimates and associated uncertainty estimates in the form of prediction intervals.

\autoref{tab:heteroscedasticity} summarizes the correlations. There is a clear tendency toward heteroscedasticity: in 10 out of 14 event logs, both the absolute prediction error and the prediction interval width are positively correlated with the true remaining time. In several logs (e.g., BPIC15-3, BPIC20ID, BPIC20PTC, BPIC20TPD), these correlations are moderate to strong for both measures, indicating that long-running cases are substantially noisier than short-running ones. 
However, this pattern is not universal. Two datasets (BPIC15-2 and Helpdesk) show negative correlations for both measures, while BPIC17W and BPIC15-4 exhibit mixed behavior. This variation indicates that both the strength and direction of heteroscedasticity differ across processes. Nevertheless, the dominant trend is a positive association between remaining time and both prediction error and uncertainty. 
Consequently, remaining time prediction is inherently noisy, with certain regions—such as the long tail—being particularly difficult to predict. This noise can arise from a wide range of sources, including resource contention, synchronization effects, batching, task handovers, and system state fluctuations~\cite{amiri2025simple,kraus2025use}.

These results indicate clear heteroscedasticity in the right tail of the remaining time distribution. Long-delay cases are not only rare, due to target imbalance, but also substantially more variable and harder to predict. Prediction difficulty in the tail, therefore, arises from both data scarcity and increased noise.

%% file: Tables/event_logs.tex
\begin{table}[ht]
\caption{Characteristics of the 14 event logs used for evaluation. Case duration is in days. $\gamma$ denotes the skewness of the remaining-time distribution.}
\label{tab_logs}
\centering
\scriptsize
\setlength{\tabcolsep}{2.5pt}
\begin{tabular}{l r r r r r : l r r r r r}
\toprule
 &  &  & \multicolumn{2}{c}{Case duration} &  & 
 &  &  & \multicolumn{2}{c}{Case duration} &  \\
\cmidrule(lr){4-5} \cmidrule(lr){10-11}
Event log & Cases & Events & Avg. & Max & $\gamma$ &
Event log & Cases & Events & Avg. & Max & $\gamma$  \\
\midrule
P2P      &   608 &   9119 &  21.5 &  109.4 & 2.8  & Helpdesk  &  4580 &  21348 &  40.9 &   60.0 &  -0.1 \\
BPIC17W  & 30276 & 240854 &  12.7 &  288.9 & 4.2  & Sepsis    &  1050 &  15214 &  28.5 &  422.3 &  3.0  \\
BPIC15-1 &  1199 &  52217 &  95.9 & 1486.0 & 5.2  & BPIC20ID  &  6449 &  72151 &  86.5 &  742.0 &  3.5  \\
BPIC15-2 &   832 &  44354 & 160.3 & 1326.0 & 3.2  & BPIC20DD  & 10500 &  56437 &  11.5 &  469.2 &  11.5 \\
BPIC15-3 &  1409 &  59681 &  62.2 & 1512.0 & 10.0 & BPIC20PTC &  2099 &  18246 &  36.8 &  328.2 &  4.0  \\
BPIC15-4 &  1053 &  47293 & 116.9 &  927.0 & 3.7  & BPIC20TPD &  7065 &  86581 &  87.4 & 1202.0 &  2.6  \\
BPIC15-5 &  1156 &  59083 &  98.0 & 1344.0 & 5.2  & BPIC20RFP &  6886 &  36796 &  12.0 &  410.0 &  10.6  \\
\bottomrule
\end{tabular}
\end{table}

%% file: Tables/heteroscedastic.tex
\begin{table}[t]
\centering
\caption{Spearman correlation between ground-truth remaining time $y$ and (i) absolute prediction error $|e|$ and (ii) prediction interval (PI) width produced by the survival model (Surv).}
\label{tab:heteroscedasticity}
\setlength{\tabcolsep}{4pt}
\begin{tabular}{l S[table-format=-1.2] S[table-format=-1.2] : l S[table-format=-1.2] S[table-format=-1.2]}
\toprule
Dataset & {$\rho(y,|e|)$}  & {$\rho(y,\text{PI}_{\text{Surv}})$} & Dataset & {$\rho(y,|e|)$} &  {$\rho(y,\text{PI}_{\text{Surv}})$}\\
\midrule
P2P      &  0.89  & 0.46  & Helpdesk & -0.41 &  -0.53\\
BPIC17W  &  0.38  & -0.43 & Sepsis   &  0.51 &  0.39\\
BPIC15-1 &  0.30  & 0.55 & BPIC20ID &  0.64 &   0.80\\
BPIC15-2 & -0.20  & -0.22 & BPIC20DD &  0.47 &   0.21\\
BPIC15-3 &  0.68  & 0.64 & BPIC20PTC & 0.56 &  0.67\\
BPIC15-4 & -0.69  & 0.09 & BPIC20TPD & 0.61 &  0.68\\
BPIC15-5 &  0.60  & 0.32 & BPIC20RFP & 0.55 &  0.22\\
\bottomrule
\end{tabular}
\end{table}

%% file: Sections/03_Setting.tex
\section{Problem Setting and Experimental Setup}
\label{sec:setting}
In this section, we define the remaining time prediction task and describe the experimental setup used to evaluate the proposed approaches.

\mypar{Problem Definition}
We formalize remaining time prediction as a supervised learning task based on event log data.

\mypartwo{Event Log, Traces, Event Prefixes}
Process execution data are recorded in an event log $\mathcal{E}$, which is a collection of traces. Each trace $\sigma = \langle e_{1}, e_{2}, ..., e_{n}\rangle$ represents the chronological sequence of events for a single process instance (case). An event is a tuple $e = (c, a, t, \Delta)$, where $c$ is the case identifier, $a$ the executed activity, $t$ the timestamp, and $\Delta$ an attribute-value map with optional data payload. We define the projection $\pi_T(e) = t$ to extract an event's timestamp. A prefix of length $m \in [1, n-1]$ is denoted $\sigma^m=\langle e_{1}, \ldots, e_{m}\rangle$ and represents partial execution up to the $m$-th event.

\mypartwo{Feature Extraction}
Remaining time prediction is formulated as a regression task, starting with feature extraction. Event prefixes of varying lengths are encoded into feature vectors $\mathbf{x} = \Gamma(\sigma^m)$, and target values $y = \pi_T(e_n) - \pi_T(e_m)$.  This yields the training dataset $\mathcal{D}=\{(\mathbf{x}_{i}, y_{i})\}_{i=1}^{N}$, where $\mathbf{x}_{i}\in\mathbb{R}^{d}$ denotes the input and $y_{i}\in\mathbb{R}$ the target.

\mypartwo{Problem Statement}
Given the training dataset $\mathcal{D}$, the goal is to approximate the function $y = f(\mathbf{x})$ that maps inputs to remaining time. The predictive model typically follows an encoder–head structure: an encoder $\mathbf{z} = g(\mathbf{x}, \theta)$ produces a latent representation $\mathbf{z}$, which is mapped to the predicted remaining time $\hat{y} = h(\mathbf{z}, \psi)$. The parameters $\theta$ and $\psi$ are optimized via supervised learning to minimize a regression loss:

\begin{equation}
\mathcal{L}(\theta, \psi) =
\begin{cases}
\frac{1}{N} \sum_{i=1}^{N} w_i \, |\hat{y}_i - y_i|, & \text{if absolute error is used}, \\[8pt]
\frac{1}{N} \sum_{i=1}^{N} w_i \, (\hat{y}_i - y_i)^2, & \text{if squared error is used.}
\end{cases}
\label{eq:loss_function}
\end{equation}

Note that standard loss functions (e.g., MAE, MSE) assign equal weight ($w_i=1$) to all training examples, biasing models toward frequent target values.



\mypar{Experimental Setup}
For the experiments conducted in the subsequent sections, we will use the following event logs, models, and metrics.
The event logs, implementation details, and configurations are available in our repository.\footnote{\url{https://zenodo.org/records/20433645}}

\mypartwo{Event Logs and Data Split}
We use 14 event logs for evaluation to ensure robust and generalizable results (see Table~\ref{tab_logs}). We apply a temporal hold-out split, dividing each log into 64\% training, 16\% validation, and 20\% test data.

\mypartwo{Model}
We use a data-aware LSTM model proposed by Navarin et al. \cite{navarin2017lstm}, which predicts remaining time from event prefixes encoded as sequences of feature vectors (control-flow, temporal, and all data attributes), left-padded to a fixed length. The model consists of two LSTM layers (hidden size 150) with layer normalization and dropout (0.1), followed by a dense linear regression head that maps the final hidden representation to the predicted remaining time. Training uses AdamW, batch size 128, early stopping (patience 30), and up to 300 epochs.

\mypartwo{Hyperparameter Tuning}
Hyperparameters are tuned on the validation set using Bayesian optimization. We define approach-specific search spaces, while keeping the model architecture fixed (see details in the supplementary repository).

\mypartwo{Evaluation Metrics}
To analyze prediction difficulty across target regions, we partition test samples into many/medium/few groups using quantile-based splits. Thresholds are computed from the training and validation targets. By default, the few-shot region corresponds to the highest 10\% of target values, the medium region to the next 30\%, and the many-shot region to the remaining 60\%.
To enable comparisons across datasets with different cycle time scales, we use a normalized mean absolute error (nMAE) \cite{roider2024assessing}. For a region $r$, it is defined as
\[
\mathrm{nMAE}_r =
\frac{\frac{1}{|r|}\sum_{i\in r} |y_i - \hat{y}_i|}
{\frac{1}{N}\sum_{i=1}^{N} |y_i - \mathrm{median}(y_{\text{train}})|},
\]
where $y_i$ and $\hat{y}_i$ denote true and predicted remaining times, $|r|$ is the number of samples in region $r$, and $N$ is the total number of test samples. 
The denominator corresponds to the mean absolute deviation from the training-set median, i.e., the error of a constant median predictor. This normalizes the MAE relative to a trivial baseline, enabling assessment of prediction difficulty and model performance across regions and datasets on a common scale.
Consequently, an nMAE of 1 reflects performance comparable to this baseline, values below 1 indicate improved predictions, and values above 1 indicate worse performance. This interpretation enables direct comparison across many-, medium-, and few-shot regions.

%% file: Sections/04_Imbalanced.tex
\section{Assessing the Effectiveness of Imbalanced Regression}
\label{sec:imbalanced}
The remaining time distribution is often highly skewed, with only a small fraction of cases exhibiting large delays. As a result, standard machine learning models tend to perform poorly on these rare cases, which naturally leads to the hypothesis that \textit{target imbalance} is a key source of the problem. In this section, we therefore evaluate imbalanced regression approaches designed to address this issue, which have shown promising results in time series forecasting, computer vision, and natural language processing~\cite{branco2017smogn,yang2021delving,silva2022model,ren2022balanced}.

\subsection{Data-level Approaches: Resampling the Training Distribution}
We describe data-level approaches for imbalanced regression and evaluate their effectiveness for remaining time prediction.

\mypar{Approach}
Data-level approaches mitigate imbalance via resampling the training data, either by \textit{undersampling} frequent target ranges or \textit{oversampling} rare ones.
Oversampling methods generate synthetic samples in sparse regions of the target space, e.g., by interpolating between neighboring observations as in SMOTER~\cite{torgo2013smote}, or by perturbing rare samples with Gaussian noise~\cite{branco2016ubl}.
Hybrid methods such as SMOGN combine both strategies by undersampling frequent regions and oversampling rare targets through interpolation and noise injection~\cite{branco2017smogn}.

Applying data-level approaches to PPM is non-trivial. Remaining time prediction operates on event prefixes encoded as left-padded sequences, where the meaning of an input depends on both feature values and their position. Naive interpolation can produce unrealistic samples by mixing padding with valid events or combining prefixes of different lengths, distorting the sequential structure and violating admissible process behavior.
We adapt SMOGN to this setting as a feature-space augmentation procedure applied after prefix encoding. Rare samples are identified using a relevance function derived from the training target distribution, assigning higher importance to large remaining times. To avoid mixing prefixes of different temporal extent, we first group samples by their actual prefix length, computed from the non-padded part of the left-padded input sequence.
Nearest neighbors are computed only within these prefix-length groups. For each prefix, the padded rows are discarded, the remaining encoded event-level feature vectors are flattened, and Euclidean distance is used to select neighbors. Thus, padding positions do not influence neighbor selection.
Synthetic samples are generated in the encoded numerical input space, rather than from raw trace symbols or learned embedding vectors. For interpolation-based oversampling, we linearly interpolate the non-padded feature vectors of a rare prefix and one of its neighbors from the same prefix-length group; the padded part remains unchanged. The remaining-time target is interpolated analogously. Alternatively, Gaussian noise is added only to the non-padded part of rare prefixes, scaled by the empirical feature-wise standard deviation within the group. Frequent samples are reduced via random undersampling, while all rare samples are retained.

\begin{figure}[t]
    \centering
    \includegraphics[width=\linewidth]{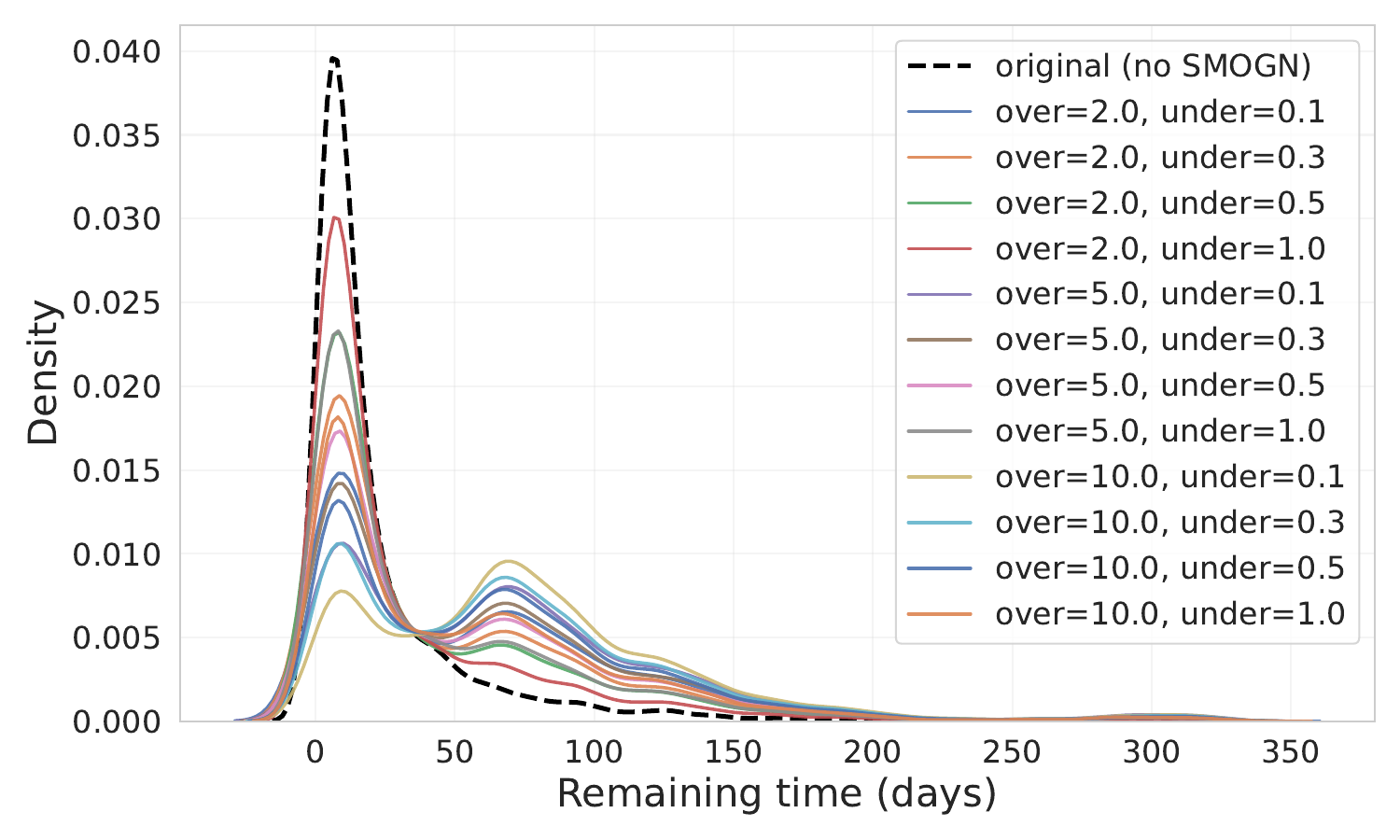}
    \caption{BPIC20PTC target distribution for various configurations of SMOGN.}
    \label{fig:SMOGN}
 \end{figure}

\mypar{Experiments}
\autoref{fig:SMOGN} shows how SMOGN reshapes the training distribution for BPIC20PTC under different configurations. We vary the oversampling factor for rare cases and the undersampling ratio for frequent ones. Increasing both shifts probability mass toward higher remaining times, increasing the density of delayed cases. However, this does not improve predictive performance: the MAE remains between 8.1 and 9.4 days across configurations.

We apply SMOGN to all 14 event logs and compare it to the same LSTM trained on the original (imbalanced) data. Averaged over five seeds per log, SMOGN reduces nMAE in the few-shot region by 8.0\% but increases error in the medium and many regions by 20.5\% and 31.4\%, respectively, resulting in a 17.5\% higher overall nMAE. Detailed results are provided in the supplementary GitHub repository.

Overall, even substantial changes to the training distribution do not improve predictive performance, suggesting that data scarcity in the tail is not the primary bottleneck. We therefore turn to algorithm-level approaches.

\subsection{Algorithm-level Approaches: Reweighting the Loss Function}
Algorithm-level approaches mitigate imbalance by modifying the regression objective in \autoref{eq:loss_function} to assign greater weight to underrepresented target regions. We first introduce these approaches and then evaluate their effectiveness for remaining time prediction.

\mypar{Approaches}
We consider cost-sensitive weighting, distribution-balanced objectives, error-aware losses, and relevance-based error measures, and refer to the original publications for details.

\mypartwo{Cost-Sensitive Weighting (CSW)}
CSW assigns static loss weights to different target ranges based on the empirical target distribution. The continuous target domain is discretized into bins, and their frequencies are estimated from the training data. Samples in less frequent bins receive higher weights, while those in dense regions are down-weighted. In practice, weights are set inversely proportional to the bin frequency or its square root~\cite{yang2021delving}.

\mypartwo{Balanced Mean Squared Error (BMSE)}
BMSE incorporates the target distribution directly into the regression objective. We use the batch-based Monte Carlo (BMC) variant proposed in~\cite{ren2022balanced}. Instead of applying static weights, BMC contrasts the prediction error of a sample with the errors obtained for other target values within the same mini-batch. This is implemented through a softmax-style normalization over squared errors across batch targets. Consequently, prediction errors associated with dense target regions receive relatively less influence than those in sparse regions. 

\mypartwo{Error-Aware Loss (EAL)}
EAL adjusts sample contributions based on prediction error magnitude. We consider the Focal-R loss~\cite{yang2021delving}, which modulates the error by a nonlinear function that increases with its magnitude. This reduces the contribution of small errors while increasing the relative influence of larger errors. As a result, the optimization focuses more on difficult examples, which are often associated with rare or extreme target values.

\mypartwo{Squared Error Relevance Area (SERA)}
SERA~\cite{silva2022model} is a relevance-based loss for imbalanced regression. It relies on a relevance function $\phi : Y \rightarrow [0,1]$ that assigns each target value an importance score, allowing non-uniform preference over the continuous target domain~\cite{torgo2007utility,branco2016ubl}. In this work, $\phi$ is automatically derived from the training target distribution using box-plot statistics~\cite{ribeiro2020imbalanced}, assigning higher relevance to extreme remaining time values. Consequently, prediction errors on highly relevant targets receive greater weight, while errors on low-relevance targets contribute less to the overall loss.

\input{Tables/statistical_test}

\mypar{Experiments}
We compare the above approaches to a baseline trained with standard MAE (Vanilla), using nMAE across different target regions averaged over five random seeds. A Friedman test reveals significant differences across approaches ($p<0.001$ for all regions: all, many, medium, few), motivating a more detailed comparison. In \autoref{tab:algorithm_level_statistical}, approaches are ranked per dataset by nMAE within each region and averaged across datasets, while significant pairwise differences are identified using the Wilcoxon signed-rank test.

In the many-shot region, Vanilla attains the best average rank (1.64), while SERA performs significantly worse than Vanilla, CSW, and EAL. In the medium-shot region, results are comparable, with SERA again underperforming relative to CSW and BMSE. A different pattern emerges in the few-shot region, where SERA achieves the best rank (1.00) and significantly outperforms Vanilla, CSW, and EAL; BMSE ranks second and also improves over Vanilla.  However, these gains do not translate to overall performance: SERA performs significantly worse overall, while EAL and CSW achieve the best ranks. 

\begin{figure}[t]
\centering
\begin{subfigure}[b]{0.49\textwidth}
    \includegraphics[width=\linewidth]{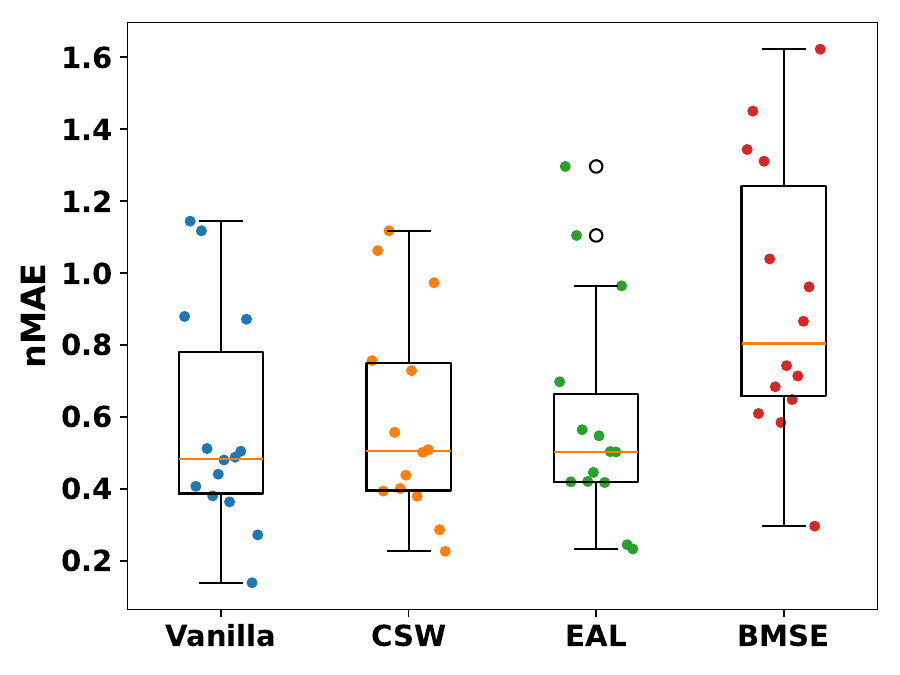}
    \caption{Many}
    \label{fig:nmae_many}
\end{subfigure}
\hfill
\begin{subfigure}[b]{0.49\textwidth}
    \includegraphics[width=\linewidth]{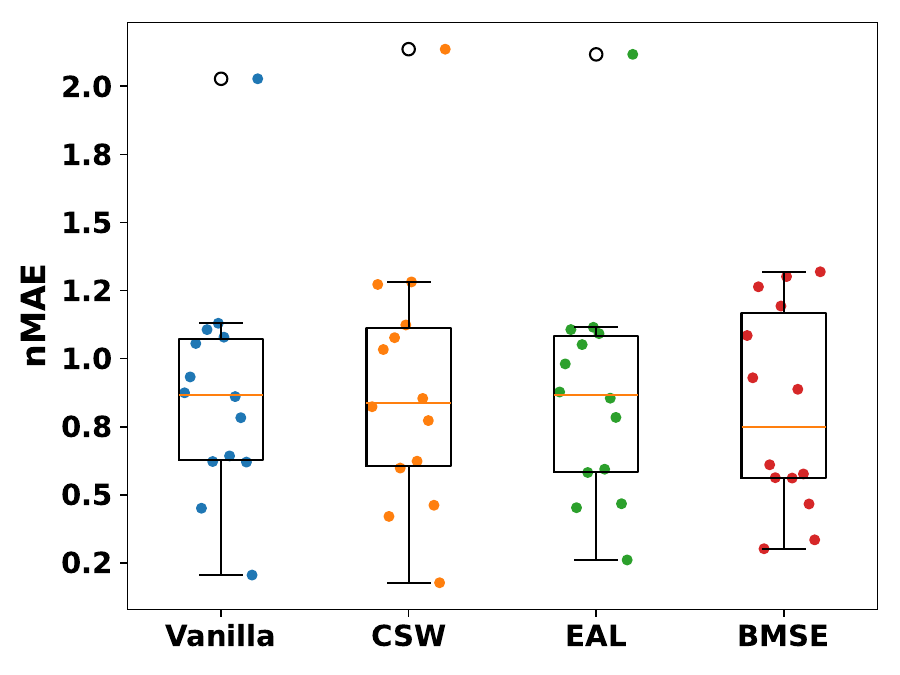}
    \caption{Medium}
    \label{fig:nmae_medium}
\end{subfigure}
\vspace{2mm}
\begin{subfigure}[b]{0.49\textwidth}
    \includegraphics[width=\linewidth]{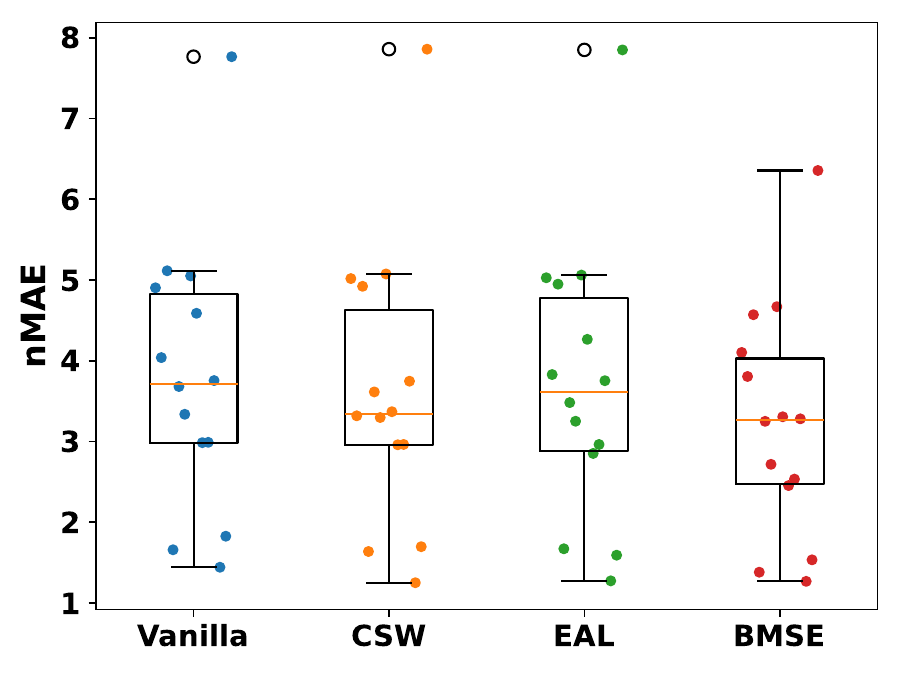}
    \caption{Few}
    \label{fig:nmae_few}
\end{subfigure}
\hfill
\begin{subfigure}[b]{0.49\textwidth}
        \includegraphics[width=\linewidth]{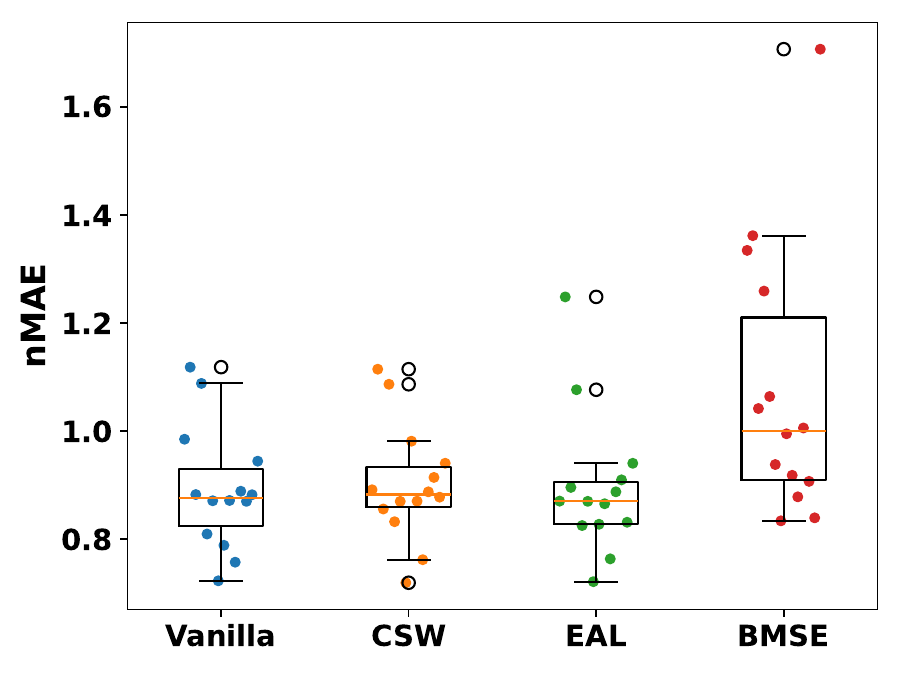}
    \caption{All}
    \label{fig:nmae_all}
\end{subfigure}
\caption{Distribution of nMAE across datasets for algorithm-level imbalanced regression approaches, analyzed across different target-frequency regions.}
\label{fig:nmae_boxplots}
\end{figure}

The boxplots in \autoref{fig:nmae_boxplots} show nMAE distributions across datasets for each approach. SERA is omitted due to its substantially larger errors in the many- and medium-shot regions and overall. In the many-shot region, Vanilla, CSW, and EAL exhibit nearly identical medians and similar variability, while BMSE shows a slightly higher median and greater spread. A similar pattern appears in the medium-shot region, with increased variability and higher errors across all approaches. In contrast, the few-shot region exhibits substantially higher variability across datasets and consistently large errors, with nMAE often exceeding 1. Since nMAE is normalized by the error of a constant median predictor, values above 1 indicate performance worse than this trivial baseline. This highlights the difficulty and instability of prediction in the long-tail region. 

We also evaluate feature and label distribution smoothing~\cite{yang2021delving} in combination with these approaches, but observe only marginal and inconsistent effects (see supplementary repository).

Overall, algorithm-level approaches do not consistently improve prediction accuracy in the long tail. When improvements occur (e.g., SERA), they come at the cost of worse performance in the many-shot region and overall accuracy. Thus, if the goal is solely to identify high-delay cases, SERA may be effective. However, in most practical settings, models are expected to perform well across all cases.

%% file: Tables/statistical_test.tex
\begin{table}[t]
  \centering
  \caption{Comparison of algorithm-level approaches using average ranks (lower is better) and significant pairwise differences identified by the Wilcoxon signed-rank test. Results are reported for nMAE across target regions.}
  \setlength{\tabcolsep}{4pt}
  \renewcommand{\arraystretch}{1.15}
  \scriptsize
  \newcolumntype{L}[1]{>{\RaggedRight\arraybackslash}p{#1}}
  \begin{tabularx}{\linewidth}{@{} l L{0.36\linewidth} L{0.52\linewidth} @{}}
    \toprule
    Mode & Average Ranks (lower is better) & Significant Performance Differences (Wilcoxon signed-rank test $p-values$)  \\
    \midrule
    Many & Vanilla (1.64), CSW (2.14), EAL (2.36), BMSE (3.86), SERA (5.00) & Vanilla > SERA (0.001); CSW > SERA (0.006); EAL > SERA (0.015) \\
    \midrule
    Medium & CSW (2.21), BMSE (2.36), EAL (2.64), Vanilla (2.86), SERA (4.93) & CSW > SERA (0.012); BMSE > SERA (0.020) \\
    \midrule
    Few & SERA (1.00), BMSE (2.14), CSW (3.57), EAL (3.79), Vanilla (4.50) & Vanilla vs BMSE (0.042); SERA > Vanilla (0.000); SERA > CSW (0.020); SERA > EAL (0.009) \\
    \midrule
    All & EAL (1.93), CSW (2.00), Vanilla (2.36), BMSE (3.71), SERA (5.00) & Vanilla > SERA (0.015); CSW > SERA (0.004); EAL > SERA (0.003) \\
    \bottomrule
  \end{tabularx}
  \label{tab:algorithm_level_statistical}
\end{table}

%% file: Sections/05_uncertainty.tex
\section{Assessing the Effectiveness of Uncertainty Modeling}
\label{sec:uncertainty}
Building on the observation from \autoref{sec:motivation} that heteroscedastic uncertainty increases with remaining time and is particularly pronounced in the long tail of the distribution, we investigate whether this relationship can be exploited to improve delay detection. To this end, we turn the previously used deterministic LSTM into a probabilistic model via survival analysis, enabling it to produce explicit uncertainty estimates. These estimates are then used as input to a downstream classifier that predicts whether a case will exceed a given delay threshold. This reformulation is not intended to replace remaining time regression, but rather to study whether uncertainty estimates can help distinguish highly delayed cases from normal executions. We first introduce the uncertainty modeling framework and then evaluate its utility for delay detection.


\mypar{Modeling Uncertainty via Survival Analysis}
To capture predictive uncertainty, we model remaining time using a discrete-time survival formulation, treating process completion as a time-to-event problem~\cite{george2014survival}. Given an input feature vector $\mathbf{x}$, instead of predicting a single remaining time value $\hat{y}$, the model estimates a distribution over future completion times. Concretely, the remaining time horizon is discretized into $K$ ordered intervals using quantile-based binning of the training targets, where $K$ is treated as a tunable hyperparameter. For each interval $k \in \{1, \dots, K\}$, the model predicts a hazard value $h_k(\mathbf{x})$, corresponding to the conditional probability that the case completes in interval $k$ given that it has not completed earlier.

The predictive model uses the same encoder as the Vanilla LSTM model to produce a latent representation $\mathbf{z} = g(\mathbf{x}, \theta)$, which is mapped to $K$ hazard logits via a dedicated output layer. These logits are transformed into probabilities representing the discrete hazard function. Training minimizes the negative log-likelihood of the observed event time under this formulation.

From the predicted hazards, the full discrete remaining time distribution is reconstructed via the corresponding survival and event probabilities. This allows deriving both point estimates and predictive uncertainty measures. In our implementation, we extract summary statistics including the expected remaining time and prediction intervals directly from this distribution. Heteroscedastic uncertainty is quantified by the width of the central 80\% prediction interval. Further implementation details are provided in the accompanying repository.


\mypar{Delay Detection via Uncertainty Modeling}
To evaluate the utility of uncertainty estimates, we formulate delay detection as a binary classification task. A case is labeled as delayed if its total duration exceeds a predefined quantile threshold (e.g., $q=0.8$) of the cycle times in the training and validation sets. The goal is to predict this label for ongoing cases based on their observed prefixes.

\begin{figure}[!htbp]
  \centering
  \includegraphics[scale=0.39]{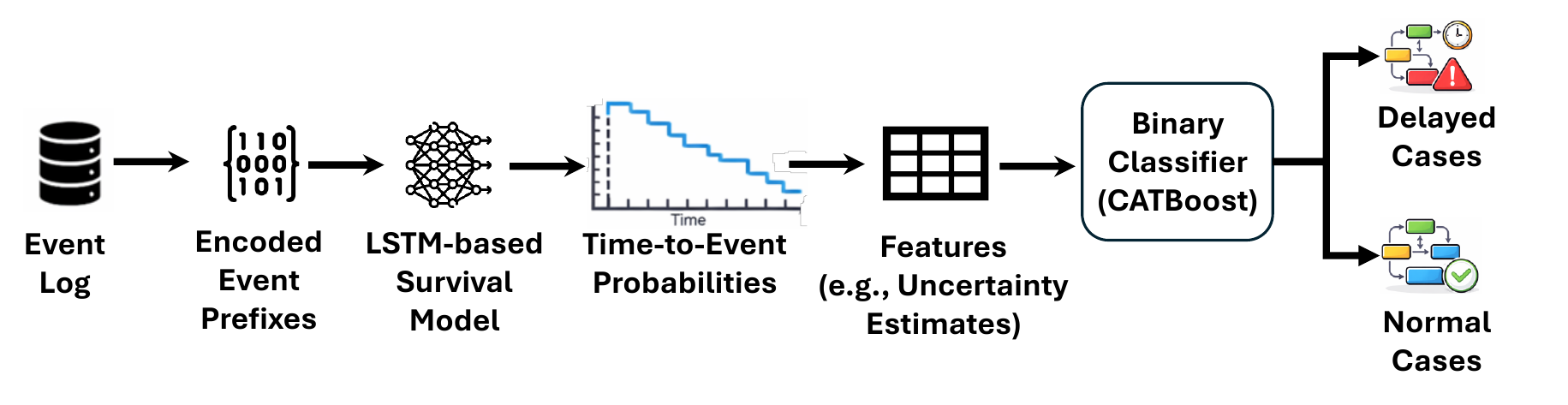}
  \caption{Uncertainty-aware delay detection pipeline using survival-based predictions and a tabular classifier.}
  \label{Approach}
\end{figure}

As illustrated in Figure~\ref{Approach}, we develop an \textit{uncertainty-aware model} that uses the outputs of the previously described survival model. Specifically, we extract distributional summaries (mean and median remaining time), uncertainty measures (prediction standard deviation and the widths of 80\% and 90\% prediction intervals), tail mass (i.e., the probability that the remaining time exceeds the modeled horizon), and temporal context features (elapsed time since the start of the case and time since the last event). These features are used as input to a gradient boosting classifier (CatBoost). This design enables the model to incorporate uncertainty signals about future completion times, rather than relying solely on prefix features.

\mypar{Results}
We evaluate the performance of our uncertainty-aware model in detecting high-delay cases using recall, precision, F1-score, and PR-AUC across all 14 event logs in \autoref{tab:performance_metrics_20}. We compare our model against a \textit{baseline}, which shares the same LSTM encoder as the survival model, but---instead of modeling uncertainty---directly produces a point estimate, classifying the prefix either as normal or high-delay. 
The baseline achieves very low recall and therefore misses most delayed cases. In contrast, the uncertainty-aware model substantially improves recall across all datasets, increasing the average from 0.21 to 0.61. This means that a much larger proportion of delayed cases can be identified in advance, which is essential for timely intervention. For instance, on the BPIC20PTC log, the baseline correctly identifies only 1 out of 175 high-delay cases, whereas the uncertainty-aware model detects 132.
A Wilcoxon signed-rank test ($p<0.0001$) confirms that this improvement is statistically significant.

\input{Tables/performance_table_20}

These gains are achieved while maintaining competitive precision: in most cases, precision remains comparable or improves (e.g., BPIC17W: 0.48 to 0.77), with no statistically significant difference overall ($p=0.17$). Consequently, the F1-score consistently improves across datasets, reflecting a better balance between precision and recall. Similarly, PR-AUC improves in nearly all cases, indicating improved ranking of delayed cases. These improvements are statistically significant (Wilcoxon test: $p<0.0001$ for F1-score, $p<0.0003$ for PR-AUC).

Overall, these results demonstrate that incorporating predictive uncertainty improves delay detection performance by substantially increasing the ability to identify delayed cases without sacrificing precision. This suggests that uncertainty provides actionable information for prioritizing high-risk cases in practice. We observe similar improvements under a stricter delay definition ($q=0.9$), as reported in the supplementary repository. However, since delay detection abstracts the original regression problem into a threshold-based classification task, these findings should not be interpreted as improvements in remaining time estimation itself.

To further assess how early delayed cases can be identified during process execution, we perform an earliness analysis that evaluates model performance at different stages of a running case. Specifically, each prefix is interpreted as a partial observation of an ongoing process instance, and its length is normalized by the total number of events in that case (i.e., its full execution trace). This allows us to group prefixes according to how far the case has progressed (e.g., early, mid, and late stages), and to evaluate performance on subsets of prefixes corresponding to up to 20\%, 40\%, 60\%, and 80\% of the process execution. The uncertainty-aware model consistently outperforms the baseline across all prefix ratios, with the most pronounced gains at early stages. In particular, recall is substantially higher even when only a small portion of the case is observed, while precision remains comparable, resulting in consistently improved F1-scores and PR-AUC. Although the performance gap narrows as more of the case becomes available, the uncertainty-aware model remains superior throughout, indicating that incorporating predictive uncertainty enables earlier and more reliable detection of delayed cases.

%% file: Tables/performance_table_20.tex
\begin{table}[t]
\centering
\caption{Delay detection performance (threshold $q=0.8$). Average Recall, Precision, F1-score, and PR-AUC over five random seeds for baseline and uncertainty-aware models.}
\label{tab:performance_metrics_20}
\footnotesize
\setlength{\tabcolsep}{2.2pt}
\begin{tabular}{lrrrrrrrr}
\toprule
& \multicolumn{4}{c}{Baseline Classifier} & \multicolumn{4}{c}{Uncertainty-Aware Model} \\
\cmidrule(lr){2-5} \cmidrule(lr){6-9}
Dataset & Recall & Precision & F1-Score & PRAUC & Recall & Precision & F1-Score & PRAUC \\
\midrule
P2P & 0.11 & \textbf{0.99} & 0.20 & 0.74 & \textbf{0.83} & 0.95 & \textbf{0.89} & \textbf{0.91} \\
BPIC17W & 0.34 & 0.48 & 0.40 & 0.38 & \textbf{0.59} & \textbf{0.77} & \textbf{0.67} & \textbf{0.59} \\
BPIC15-1 & 0.49 & 0.34 & 0.39 & 0.23 & \textbf{0.66} & \textbf{0.40} & \textbf{0.49} & \textbf{0.30} \\
BPIC15-2 & 0.38 & 0.11 & 0.17 & 0.07 & \textbf{0.75} & \textbf{0.12} & \textbf{0.21} & \textbf{0.10} \\
BPIC15-3 & 0.31 & 0.39 & 0.34 & 0.23 & \textbf{0.65} & \textbf{0.65} & \textbf{0.65} & \textbf{0.48} \\
BPIC15-4 & 0.40 & 0.04 & 0.07 & 0.04 & \textbf{0.75} & \textbf{0.06} & \textbf{0.12} & \textbf{0.05} \\
BPIC15-5 & 0.33 & \textbf{0.42} & 0.36 & 0.24 & \textbf{0.53} & 0.41 & \textbf{0.46} & \textbf{0.29} \\
HelpDesk & 0.02 & \textbf{0.32} & 0.04 & 0.21 & \textbf{0.66} & 0.25 & \textbf{0.36} & \textbf{0.24} \\
Sepsis & 0.19 & \textbf{0.59} & 0.27 & \textbf{0.26} & \textbf{0.51} & 0.28 & \textbf{0.36} & 0.24 \\
BPIC20ID & 0.05 & 0.04 & 0.04 & 0.02 & \textbf{0.39} & \textbf{0.12} & \textbf{0.18} & \textbf{0.06} \\
BPIC20DD & 0.11 & 0.52 & 0.18 & 0.30 & \textbf{0.47} & \textbf{0.65} & \textbf{0.55} & \textbf{0.45} \\
BPIC20PTC & 0.01 & 0.18 & 0.01 & 0.06 & \textbf{0.76} & \textbf{0.20} & \textbf{0.31} & \textbf{0.16} \\
BPIC20TPD & 0.07 & 0.22 & 0.11 & 0.11 & \textbf{0.46} & \textbf{0.28} & \textbf{0.35} & \textbf{0.18} \\
BPIC20RFP & 0.18 & 0.73 & 0.29 & 0.42 & \textbf{0.48} & \textbf{0.74} & \textbf{0.58} & \textbf{0.53} \\
\midrule
Average & 0.21 & 0.38 & 0.21 & 0.23 & \textbf{0.61} & \textbf{0.42} & \textbf{0.44} & \textbf{0.33} \\
\bottomrule
\end{tabular}
\end{table}

%% file: Sections/06_Related_work.tex
\section{Prior Work}
\label{sec:related_work}
In this section, we review prior work on imbalanced learning in predictive process monitoring (PPM) and discuss emerging uncertainty-aware approaches for remaining time prediction.

Data imbalance is a recurring challenge across a wide range of PPM tasks~\cite{neu2022systematic}, including next activity prediction, outcome prediction, and anomaly detection. In these settings, imbalance arises from skewed class distributions, where rare but operationally critical events (e.g., exceptional activities, negative outcomes, or anomalies) are underrepresented. Existing work has addressed imbalance in PPM primarily using techniques from imbalanced classification. 

In next activity prediction, cost-sensitive learning and focal-style losses have been proposed to emphasize hard and minority examples, improving performance on infrequent activities~\cite{kappel2021cost,nguyen2020time,He2025}. At the data level, oversampling techniques have been used to rebalance class distributions~\cite{mehdiyev2020novel}. In outcome prediction, resampling approaches such as SMOTE and its variants are commonly applied to mitigate class imbalance~\cite{teinemaa2016predictive,marquez2017run}. Similar strategies have also been adopted in domain-specific applications, including loan default and payment delay prediction~\cite{zhang2025data,zeng2008using,tater2018prediction}. Finally, in process anomaly and deviation detection, where imbalance is inherent, prior work often combines representation learning with biased sampling or under-sampling to better detect rare behaviors~\cite{elaziz2023deep,grohs2023business}.

Overall, these approaches can be broadly categorized into data-level methods (e.g., resampling~\cite{marquez2017run}), algorithm-level methods (e.g., cost-sensitive learning~\cite{kappel2021cost}), and hybrid combinations~\cite{zhang2025data}. However, all existing work focuses on \emph{classification} settings. In contrast, imbalance in \emph{regression} tasks—such as remaining time prediction—has not been systematically addressed, despite the strongly skewed and long-tailed nature of target distributions in this domain.

In parallel, a small body of work has begun to investigate uncertainty-aware modeling for remaining time prediction~\cite{weytjens_learning_2022,amiri2025simple}. These approaches aim to quantify predictive uncertainty, which can provide additional information beyond point estimates. However, they do not explicitly address imbalance, nor do they study the interaction between skewed target distributions and predictive uncertainty. 

In this work, we address these gaps by systematically studying imbalanced regression in PPM, considering both data-level and algorithm-level approaches, and by investigating uncertainty-aware modeling as a complementary perspective for delay detection.

%% file: Sections/07_conclusion.tex
\section{Conclusion}
\label{sec:conclusion}
In this paper, we investigated the intrinsic difficulty of delay detection in PPM. Our analysis showed that remaining times in business processes are typically strongly right-skewed, with existing models capturing normal cases well but struggling on the rare and operationally critical high-delay cases. We further found that predictive uncertainty increases with delay magnitude, revealing pronounced heteroscedasticity in the tail of the distribution. Thus, our findings identified \textit{target imbalance} and \textit{noise} as potential sources of difficulty. While established imbalanced regression approaches yielded only limited improvements, leveraging predictive uncertainty substantially enhanced the identification of delayed cases. These findings highlight the long tail of remaining time prediction as a key challenge for PPM and identify uncertainty-aware modeling as a promising avenue for future work.

This work has several limitations that open up avenues for future research.
First, while we show that predictive uncertainty can substantially improve delay detection, the underlying sources of this uncertainty are not yet fully disentangled and may stem from multiple factors, such as process variability, missing context, or concept drift. Future work could therefore investigate richer uncertainty modeling approaches, integrate additional contextual information, and study how delay detection methods can be adapted online in changing process environments.
Second, reformulating remaining time prediction as a binary delay detection task entails a loss of information and dependence on a threshold choice. A promising direction for future work is therefore to move beyond binary classification and instead model the full conditional distribution of remaining time, enabling delay detection and related problems (e.g., service level agreement violations) to be formulated as downstream decision rules rather than fixed thresholds.
Third, our empirical evaluation is based on a set of publicly available event logs and LSTM-based predictive models. Although this reveals consistent patterns, it remains an open question how strongly these findings generalize to other process domains, industrial settings, and other architectures.